\documentclass{article}
\pdfoutput=1

\usepackage[preprint, nonatbib]{neurips_2022}

\usepackage[utf8]{inputenc} 
\usepackage[T1]{fontenc}    
\usepackage{hyperref}       
\usepackage{url}            
\usepackage{booktabs}       
\usepackage{amsfonts}       
\usepackage{nicefrac}       
\usepackage{microtype}      
\usepackage{xcolor}         

\usepackage{graphicx}
\usepackage{amsmath}

\title{A Novel Approach for Auto-Formulation of Optimization Problems}

\author{%
  Yuting Ning\textsuperscript{\rm 1,\rm 2},
  Jiayu Liu\textsuperscript{\rm 1,\rm 2},
  Longhu Qin\textsuperscript{\rm 1,\rm 2},
  Tong Xiao\textsuperscript{\rm 1,\rm 2},
  Shangzi Xue\textsuperscript{\rm 1,\rm 2}, \\
  \textbf{Zhenya Huang\textsuperscript{\rm 1,\rm 2},
  Qi Liu\textsuperscript{\rm 1,\rm 2},
  Enhong Chen\textsuperscript{\rm 1,\rm 2},
  Jinze Wu\textsuperscript{\rm 2,\rm 3}}\\
  \textsuperscript{\rm 1} Anhui Province Key Laboratory of Big Data Analysis and Application,\\
    School of Computer Science and Technology, University of Science and Technology of China\\
    \textsuperscript{\rm 2} State Key Laboratory of Cognitive Intelligence\\
    \textsuperscript{\rm 3} iFLYTEK AI Research, iFLYTEK Co., Ltd. \\
  \texttt{\{ningyt, jy251198, qlonghu, xt20020109, xueshangzi, hxwjz\}@mail.ustc.edu.cn}\\
  \texttt{\{huangzhy, qiliuql, cheneh\}@ustc.edu.cn}
}

\begin{document}

\maketitle

\begin{abstract}

  In the Natural Language for Optimization (NL4Opt) NeurIPS 2022 competition\footnote{https://nl4opt.github.io}, competitors focus on improving the accessibility and usability of optimization solvers, with the aim of subtask 1: recognizing the semantic entities that correspond to the components of the optimization problem; subtask 2: generating formulations for the optimization problem.
  In this paper, we present the solution of our team. First, we treat subtask 1 as a named entity recognition (NER) problem with the solution pipeline including pre-processing methods, adversarial training, post-processing methods and ensemble learning. Besides, we treat subtask 2 as a generation problem with the solution pipeline including specially designed prompts, adversarial training, post-processing methods and ensemble learning.
  Our proposed methods have achieved the F1-score of 0.931 in subtask 1 and the accuracy of 0.867 in subtask 2, which won the fourth and third places respectively in this competition. Our code is available at https://github.com/bigdata-ustc/nl4opt.
\end{abstract}

\section{Introduction}
Operation research, which focuses on formulating and solving real-world decision-making problems as mathematical optimization problems, has attracted much attention in recent years \cite{gupta2022artificial}. Many operation research studies have achieved impressive performances in various scenarios, such as bike-share management \cite{Beairsto2021Identifying, Ma2016Research} and revenue maximization \cite{Bitran2003AnOverview}. To solve the optimization problem automatically, models should have the capability of understanding the natural mathematical problem and generating the mathematical formulation, which has been widely studied \cite{ning2023towards, lin2021hms, huang2020neural, liu2022cognitive}.

The Natural Language for Optimization (NL4Opt) NeurIPS 2022 competition aims to leverage natural language processing techniques to convert linear programming problems to canonical formulations that solvers can understand. Along this line, there are two challenges, which correspond to two subtasks in this competition. First, how to detect problem entities from the problem description. Second, how to generate a precise meaning representation of the optimization formulation.

In this paper, we present our solutions for the above two challenges. First, for subtask 1, we treat it as a named entity recognition problem. We propose several pre-processing and post-processing rules, train the model with adversarial training and integrate different models to get the final prediction. Second, for subtask 2, following \cite{Ramamonjison2022AugmentingOR}, we treat it as a generation problem. We propose a novel prompt-guided generation framework, which leverages the named entities in problem descriptions to construct two versions of prompts. We also use adversarial training when training the generation model, and integrate different models to get the final results, which are then post-processed. Our proposed methods have achieved the F1-score of 0.931 in subtask 1 and the accuracy of 0.867 in subtask 2, which won the fourth and third places respectively in the NL4Opt competition.

Our work is organized as follows. In Section 2, we first introduce the details of the dataset and tasks of the competition. In Section 3 and Section 4, we detail our proposed methods for subtask 1 and subtask 2 respectively. In Section 5, we present the experimental results of our methods. In Section 6, we make a brief summary.

\section{Challenge}

\subsection{Dataset}

\begin{table}
  \caption{6 types of entities.}
  \label{tab:entity}
  \centering
  \begin{tabular}{lll}
    \toprule
      Entity Type   & Description     & Example \\
    \midrule
    VAR & variable name & \textit{cleaners} in \textit{A hotel employs cleaners and receptionists.} \\
    PARAM & parameter & \textit{\$350} in \textit{receptionists earn \$350 per week.}    \\
    LIMIT     & constraint limit & \textit{100} in \textit{The hotel requires a minimum of 100 workers.}     \\
    CONST\_DIR & constraint direction & \textit{minimum} in \textit{The hotel requires a minimum of 100 workers.} \\
    OBJ\_DIR & objective direction & \textit{minimize} in \textit{Formulate an LP to minimize the wage bill.} \\
    OBJ\_NAME & objective name & \textit{the wage bill} in \textit{Formulate an LP to minimize the wage bill.} \\
    \bottomrule
  \end{tabular}
\end{table}

\begin{table}
  \caption{The statistics of the dataset.}
  \label{tab:datase}
  \centering
  \begin{tabular}{lll}
    \toprule
      Statistics   & Train     & Dev \\
    \midrule
    Number of Problems & 713 & 99 \\
    Average Number of Constraints & 2.78  & 2.94     \\
    Average Number of Objectives     & 1.0 & 1.0      \\
    Average Number of CONST\_DIR & 2.27 & 2.57 \\
    Average Number of LIMIT & 2.53 & 2.62 \\
    Average Number of OBJ\_DIR & 1.0 & 1.0 \\
    Average Number of OBJ\_NAME & 2.96 & 2.86 \\
    \bottomrule
  \end{tabular}
\end{table}

The dataset of NL4Opt competition, i.e., LPWP dataset \cite{Ramamonjison2022AugmentingOR}, contains 1101 linear programming (LP) problems, of which the train, dev and test splits contain 713, 99 and 289 respectively. The training split only contains samples from the source domain (advertising, investment, sales) whereas the dev and test splits also contain samples from target domains (production, science, transportation). Each problem is provided with its problem description, labeled named entities, the order mapping of variable mentions and labeled constrained declarations. 

Table \ref{tab:entity} shows the 6 types of entities and Table \ref{tab:datase} shows some statistics of the dataset.

\subsection{Task Description}
Subtask 1 is a named entity recognition task, which aims to recognize the label of semantic entities that correspond to the components of the optimization problem. In this task, given the problem description in natural language, we expect the model to tag the entities of the optimization problems such as the objective name and constraint limits.

Subtask 2 is a formulation generation task, which targets generating the canonical formulation to solve the optimization problem. In this task, we can take the problem description in natural language, labeled entities (i.e., the output in subtask 1) and the order mapping of variable mentions as input. The output is expected to be the canonical form of optimization formulation, including constraints and objectives.

\section{Named Entity Recognition}

In this section, we introduce our solution to subtask 1: named entity recognition (NER).

\begin{figure}[t]
    \centering
    \includegraphics[width=\textwidth]{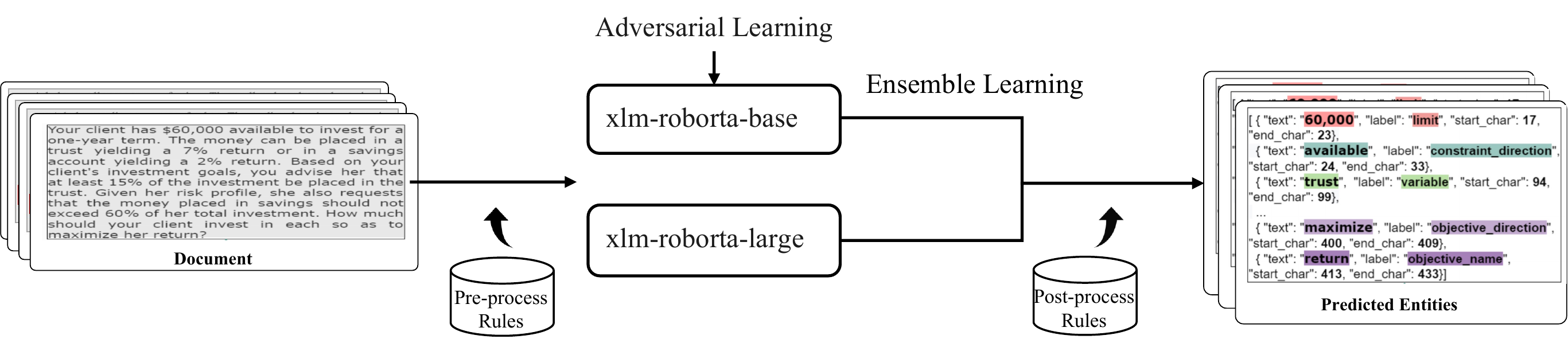}
    \caption{The proposed framework for subtask 1: named entity recognition.}
    \label{fig:task1}
\end{figure}

\subsection{Overview}
Figure~\ref{fig:task1} shows the framework of our proposed method. We build our NER system based on the pre-trained Transformer models. Specifically, given a problem description, we will first pre-process it and then feed it into the pre-trained language model. The output token representations of the input are fed into a conditional random field (CRF) layer and the CRF layer produces the label predictions. Given the label predictions of multiple models with different configurations, the ensemble module uses a category-based strategy to generate the final predictions, which will then be post-processed.

\subsection{Pre-processing}

We adopt data augmentations to expand the training set. The augmentation strategies we use are: 1) swap the variables in the problem description; 2) replace the entity tokens of objective names with their synonyms; 3) randomly replace the numbers from 0 to 9.

Besides, we notice there exists a label inconsistency in the competition data, i.e., the word \textit{times} after numbers is sometimes regarded as a part of the parameter and limit entity, but in most cases, it is not. Therefore, we perform a label standardization, i.e., removing the word \textit{times} after numbers in parameter and limit entities.

\subsection{Model}
\begin{table}
  \caption{Configurations of the selected models for ensemble learning}
  \label{tab:conf}
  \centering
  \begin{tabular}{ccccc}
    \toprule
      Model   & Language Model  &  Adversarial Training  &  Max length &  Ensemble Entity class\\
    \midrule
     1 & xlm-roberta-base & FGM  & 200 & OBJ\_DIR \\
     2 & xlm-roberta-base & PGD  & 200 & VAR \\
     3 & xlm-roberta-base & PGD  & 250 & OBJ\_NAME\\
     4 & xlm-roberta-large & PGD  & 200 & CONST\_DIR, LIMIT, PARAM \\
    \bottomrule
  \end{tabular}
\end{table}

We use XLM-RoBERTa (XLM-R) \cite{conneau2020unsupervised} as our base model. During fine-tuning, we adopt adversarial training to improve the robustness and generalization ability. Here we use the projected gradient descent method \cite{madry2017towards} and the fast gradient method \cite{miyato2016adversarial}, which could consistently improve our performances.
We train different models with different configurations and then each category of entity is predicted by the model that performs best for this category. Table \ref{tab:conf} shows the configuration details of the selected models for ensemble learning.

\subsection{Post-processing}

\begin{table}[t]
\centering
\caption{Examples of different post-processing rules in subtask 1. For short, the CD represents CONST\_DIR, ON represents OBJ\_NAME, P represents PARAM, and V represents VAR. }
  \label{tab:post}
  \resizebox{\linewidth}{!}{
  \begin{tabular}{lp{9cm}p{3cm}p{4cm}}
    \toprule
      Rule & Problem Description  & Predicted Labels & Labels after Post-processing \\
    \midrule
    1 & ... Regulars like the coffee and at least \textbf{thirty percent} of drinks must be coffee ... & B-P \textbf{O} & B-P \textbf{I-P} \\
    \midrule
    2 & ... How many of each machine should be kept in the arcade to minimize the \textbf{total number of machines} in the arcade? & \textbf{O O O} B-ON & \textbf{B-ON B-ON B-ON} I-ON \\
    \midrule
    3 & ... the number of cars used \textbf{must exceed} the number of trucks used ... & \textbf{O} B-CD & \textbf{B-CD} I-CD \\
    \midrule
    4 & ... A fitness trainer is scheduling her client to eat \textbf{more fruits and vegetables} to meet their vitamin and fiber requirements ... & \textbf{B-CD} B-V O B-V & \textbf{O} B-V O B-V \\
    
    \bottomrule
  \end{tabular}}

\end{table}

After obtaining the predicted labels with our models, we also design several post-processing rules as follows. Table \ref{tab:post} shows the examples of each rule.

\paragraph{Rule 1.} 
If the word ``times'', ``percent'' or ``\%'' is after a number labeled as PARAM or LIMIT, then the word ``times'', ``percent'' or ``\%'' will be forced to be part of the parameter or limit entity.

\paragraph{Rule 2.} If the phrase ``total number/amount/units of ... '' is before an OBJ\_NAME entity, then it will be forced to be part of the OBJ\_NAME entity.

\paragraph{Rule 3.} If the word ``must'' or the phrase ``must be'' is before a CONST\_DIR entity or a PARAM entity, then it will be forced to be a part of the CONST\_DIR entity.

\paragraph{Rule 4.} If the word ``more'' is before a VAR entity and labeled as CONST\_DIR, without other nearby CONST\_DIR entities, then it will be removed as a fake entity.

\section{Formulation Generation}

In this section, we present our solution to subtask 2: generating the precise meaning representation.

\subsection{Overview}

\begin{figure}[t]
    \centering
    \includegraphics[width=\textwidth]{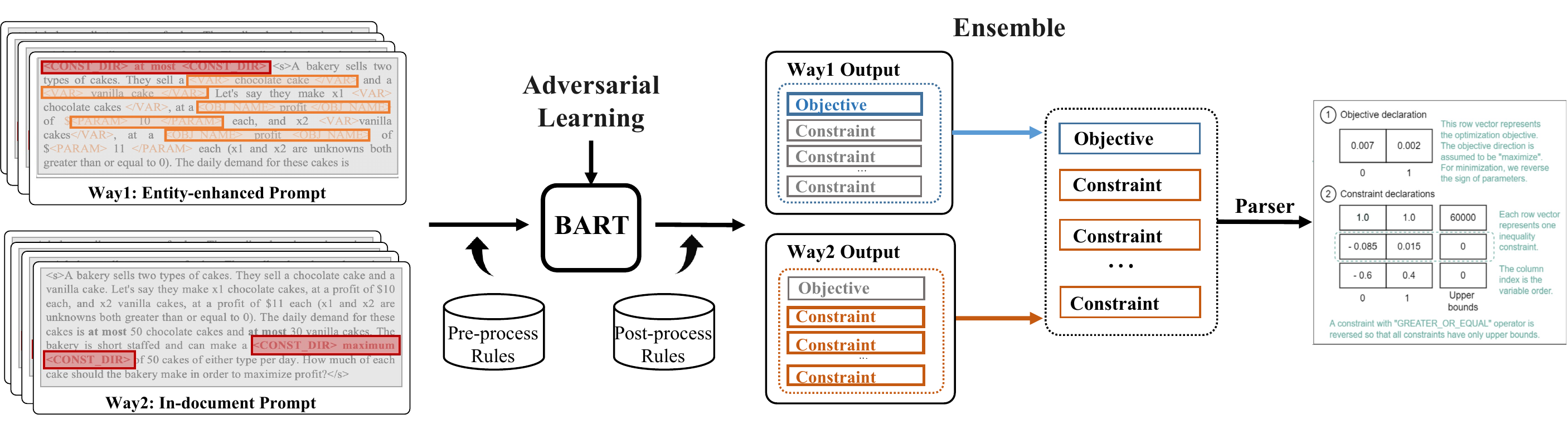}
    \caption{The proposed framework for subtask 2: formulation generation.}
    \label{fig:task2}
\end{figure}

\begin{figure}[t]
    \centering
    \includegraphics[width=\textwidth]{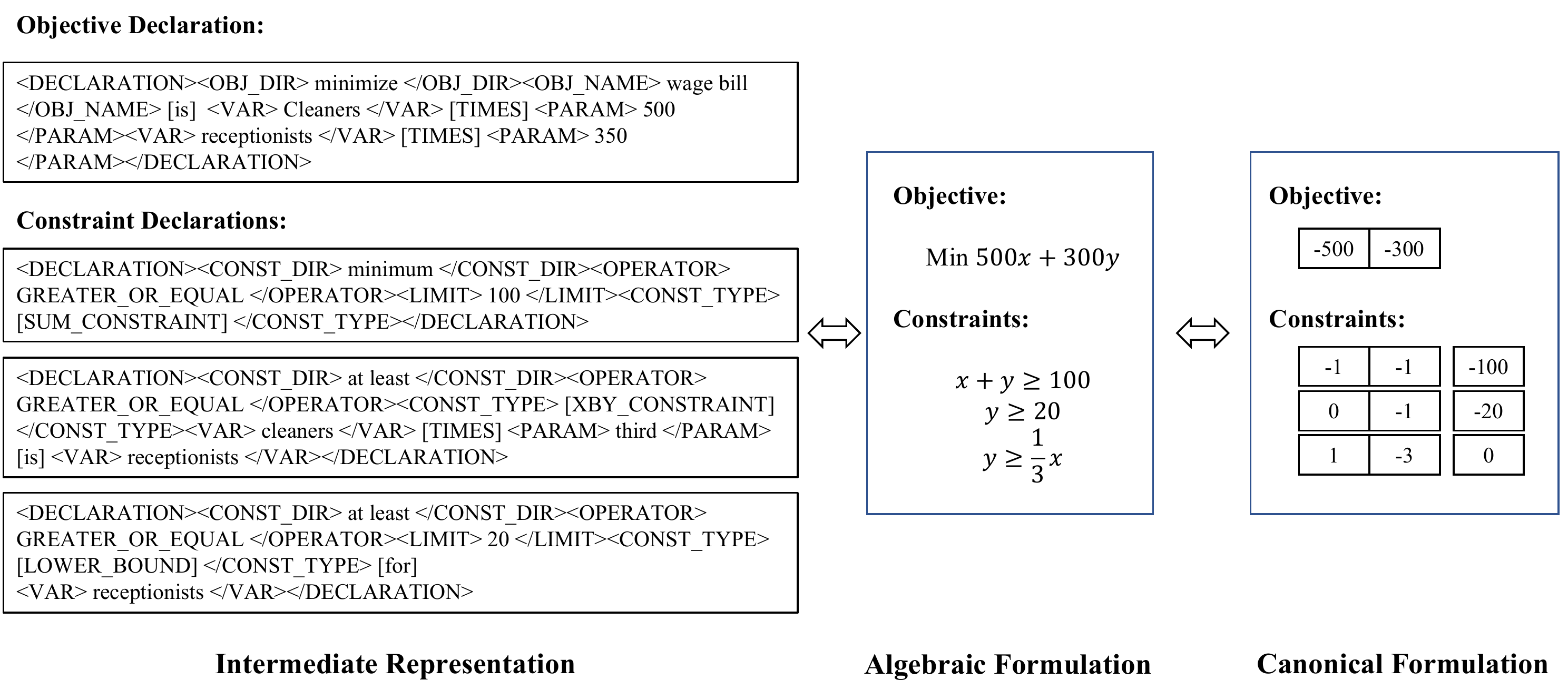}
    \caption{An example of intermediate representation, its corresponding algebraic formulation and canonical formulation.}
    \label{fig:IR}
\end{figure}

Figure~\ref{fig:task2} shows the framework of our proposed method. We use a two-stage approach for formulation generation: 1) mapping the natural language problem description to an intermediate representation (IR); 2) converting the intermediate representation to the canonical formulation with an IR parser. We focus on the first stage and propose our novel prompt-guided generation framework, using the same intermediate representation and parser as the baseline. Figure \ref{fig:IR} shows an example of the XML-looking IR declarations, corresponding algebraic formulation and canonical formulation.

The idea of our framework is to generate IR declarations one by one by using a declaration prompt. Specifically, we design two efficient versions of prompt-guided input which leverage problem description and named entities. A pre-trained encoder-decoder model is adopted to generate the IR declaration. We integrate two versions of models to generate two types of IR declarations respectively, i.e. objectives and constraints, and then propose several post-processing rules to get the final predictions. Last, the IR parser will convert IR declarations to canonical formulations. 

\subsection{Prompt-guided input}

\begin{figure}[t]
    \centering
    \includegraphics[width=\textwidth]{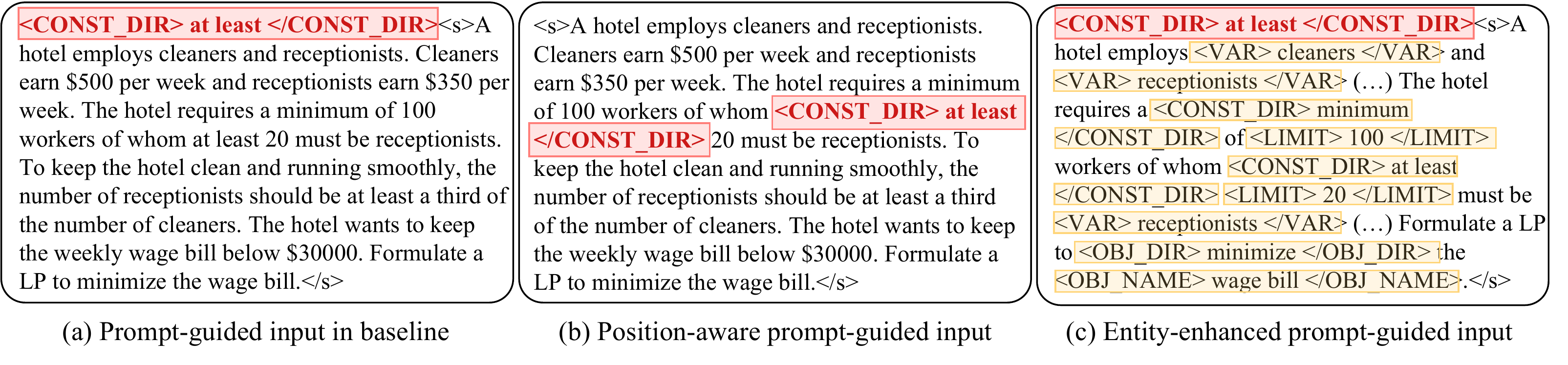}
    \caption{The examples of the prompt-guided input in the baseline and our approach.}
    \label{fig:input}
\end{figure}

The original prompt of baseline uses the entity tokens of objective direction or constraint direction as the prefix prompts, followed by the problem description. However, multiple constraint directions may have the same entity tokens but different positions, which will cause the ambiguity that different constraint declarations may have totally the same input. For example, in Figure \ref{fig:input}(a), there are two \textit{at least} corresponding to two different constraints in the problem description, which confuses the model about which one is related to the prefix prompt. Besides, the provided named entities, which carry important information, are ignored by the baseline.

To this end, in our approach, we re-design the prompt-guided input and propose two efficient versions. Figure~\ref{fig:input} shows the example of the prompt-guided input in the baseline and our approach.

\paragraph{Position-aware prompt.}
Instead of using the tokens of objective direction or constraint direction as the prefix prompt (Figure \ref{fig:input}(a)), we insert the prompt to the position of the direction entities in the document as shown in Figure \ref{fig:input}(b), which introduces the position information of the direction entities.

\paragraph{Entity-enhanced prompt.}
To leverage the rich information in named entities, as shown in Figure \ref{fig:input}(c) we mark all entities in the problem description (yellow in Figure \ref{fig:input}(c)) and add the tokens of objective direction or constraint direction as the prefix prompt (red in Figure \ref{fig:input}(c)).

\begin{figure}[t]
    \centering
    \includegraphics[width=\textwidth]{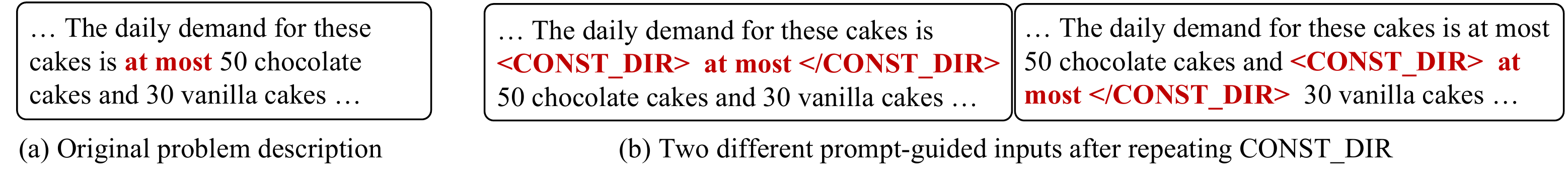}
    \caption{An example of repeating CONST\_DIR.}
    \label{fig:repeat}
\end{figure}

Besides, we notice that as we generate the input with constraint direction entities, we will get one input sample for one constraint direction entity, which will result in one predicted constraint declaration. However, when one constraint direction is followed by two limit entities, the constraint direction simultaneously constrains two variables and corresponds to two constraint declarations actually. For example, in Figure \ref{fig:repeat}(a), the constraint direction \textit{at most} limits the number of both chocolate cakes and vanilla cakes, which should be formulated by two constraint declarations. We hope to find a solution to handle this problem but has little impact on the original input. Therefore, we simply repeat the constraint direction entity before the second limit entity. Then we can get two different prompt-guided inputs and two predicted constraint declarations as shown in Figure \ref{fig:repeat}(b), which correspond to two gold declarations respectively.

\subsection{Model}
We use BART-base \cite{lewis2020bart} with copy mechanism \cite{see2017get} as our generator, which saves computing resources while achieving competitive performance compared with BART-large in other top solutions. We also adopt adversarial training to improve the robustness and generalization ability when fine-tuning. Here we use the fast gradient method, which applies perturbations to the word embeddings \cite{miyato2016adversarial}.

As we propose two versions of prompt-guided input, we fine-tune and get two models respectively. From experimental results, we notice that these two models perform well on two types of declarations. Therefore, we simply combine these two models. We use the model with the entity-enhanced prompt to predict objectives and the model with the position-aware prompt to predict constraints.

\subsection{Post-processing}

Given the generated IR declarations, we propose two post-processing rules.

\paragraph{Rule 1.} We notice there will be some mistakes in operator directions.
For example, the constraint direction is \textit{larger than}, which is obviously related to a GREATER\_OR\_EQUAL operator, but the predicted operator is LESS\_OR\_EQUAL. So the first post-processing rule is to correct wrong operators with explicit opposite constraint directions, such as \textit{larger than}.

\paragraph{Rule 2.} We also notice some wrong unseen numbers in the predicted results. For example, the predicted limit is 6500, but there is only 65000 existed in the original problem description. It is easy to notice there exists a mistake that the true number should be 65000. So the second post-processing rule is to find these wrong unseen numbers and modify them to the most similar number in the original problem description.

\section{Experiments}

\subsection{Experimental results}

\begin{table}
  \caption{Main results of our subtask 1 solution. DA: Data Augmentation. LS: Label Standardization. AT: Adversarial Training. PP: Post Processing. TEST\_F1: the F1-score on the test set.TEST\_F1$_{\text{final}}$: the F1-score on the final leaderboard.}
  \label{tab:task1}
  \centering
  \begin{tabular}{lllllll}
    \toprule
    Language Model &  DA+LS & AT & PP & TEST\_F1 & TEST\_F1$_{\text{final}}$\\
    \midrule
    xlm-roberta-base & N & N & N & 0.9160 & - \\
    xlm-roberta-base & Y & N & N & 0.9163 & - \\
    \midrule
    xlm-roberta-base & Y & FGM & N & 0.9310 & - \\
    xlm-roberta-base & Y & FGM & Y & 0.9356 & - \\
    xlm-roberta-base & Y & PGD & Y & 0.9317 & - \\
    xlm-roberta-base & Y & PGD & Y & 0.9364 & - \\
    \midrule
    xlm-roberta-large & Y & PGD & Y & 0.9358 & - \\
    \midrule
    Ensemble & & & & 0.9453 & 0.931 \\
    \bottomrule
  \end{tabular}
\end{table}

 \begin{table}
  \caption{Main experimental results of our subtask 2 solution. VAL\_ACC: the accuracy on the dev set. CONST\_ACC: the accuracy of constraints on the dev set. OBJ\_ACC: the accuracy of objectives on the dev set. TEST\_ACC: the accuracy on the test set. "*" indicates the model used for the ensemble.}
  \label{tab:task2}
  \centering
  \begin{tabular}{lllll}
    \toprule
      Method   & VAL\_ACC & CONST\_ACC & OBJ\_ACC  & TEST\_ACC \\
    \midrule
    Baseline & 0.6149 & 0.5502 & 0.7778 & 0.641 \\
    \midrule
    Position-aware prompt & 0.7586 & 0.7149 & 0.8687 & 0.803 \\
    +fgm & 0.7845 & - & - & 0.825 \\
    +repeat CONST\_DIR* & 0.8649 & - & - & 0.856\\
    \midrule
    Entity-enhanced prompt* & 0.6667 & 0.5542 & 0.9394 & 0.708 \\
    +fgm & 0.6810 & - & - & 0.693 \\
    \midrule
    Ensemble & - & - & - & - \\
    \textbf{+post-processing} & - & - & - & \textbf{0.867} \\
    \bottomrule
  \end{tabular}
\end{table}

Our final model achieves an F1-score of 0.931 on the test set in subtask 1 and an accuracy of 0.867 on the test set in subtask 2.

Table \ref{tab:task1} shows the main results of our solution for subtask 1. We notice that the major improvements of the final result come from adversarial training and model ensemble, while pre-processing and post-processing improve the performance of the model slightly.

Table \ref{tab:task2} shows the main results of our solution for subtask 2. We observe that specially designed prompts significantly improve the performances, which demonstrates that it is necessary to introduce entities' positions and other entity information. It is worth noting that the model based on entity-enhanced prompt performs well on objectives but fails on constraints. The possible reason may be that each problem has only one objective but many constraints. Though we introduce lots of entity information, the model cannot decide which parameter or limit is associated with the current constraint and thus cannot improve the accuracy of constraints.
Besides, repeating constraint directions which are followed by two LIMIT entities alleviates the mismatch between the number of the ground truth and predicted declarations.
 Moreover, it is helpful to leverage different models to predict different types of declarations.

\section{Conclusion}

In this paper, we presented the solutions of our team for the NL4Opt competition. First, we treated subtask 1 as a named entity recognition (NER) problem, solving it with pre-processing methods, adversarial training, post-processing methods and ensemble learning. Besides, we treated subtask 2 as a generation problem, solving it with specially designed prompt-guided input, adversarial training, post-processing methods and ensemble learning. We believe that our methods still have a lot of room for improvement. We hope to further research on the auto-formulation task in the future.

\bibliographystyle{plain}
\bibliography{main}

\end{document}